 \documentclass[final,5p,times,twocolumn]{elsarticle}

\usepackage{framed} 

\usepackage{multicol} 
\usepackage{nomencl} 

\makenomenclature

\renewcommand*\nompreamble{\begin{multicols}{2}}

\renewcommand*\nompostamble{\end{multicols}}

\usepackage{amssymb}
\usepackage{nomencl} 
\usepackage{amsmath}
\usepackage{multirow}

\usepackage{blindtext}
\usepackage{graphicx}
\usepackage[ruled,vlined,commentsnumbered,linesnumbered,lined,boxed]{algorithm2e}
\usepackage{algorithmic}

\usepackage{tikz}
\usepackage{siunitx}

\usepackage{tikz}
\usepackage{lettrine}
\allowdisplaybreaks
\usepackage[export]{adjustbox}
\usepackage{ragged2e}
\usepackage{subfig}
\usepackage{float}
\usepackage{booktabs}
\usepackage{makecell}  
\usepackage{tabularx}
\usepackage{siunitx}

\begin{document}

\begin{frontmatter}

\title{Optimal Uncertainty-guided Neural Network Training}



\author{H M Dipu Kabir$^a$, Abbas Khosravi$^a$, Abdollah Kavousi-Fard$^b$, Saeid Nahavandi$^a$, Dipti Srinivasan$^c$}
\address{$^a$Institute for Intelligent Systems Research and Innovation (IISRI), Deakin University, Australia.} 
\address{$^b$Department of Electrical and Electronics Engineering, Shiraz University of Technology, Iran.}
\address{$^c$Department of Electrical \& Computer Engineering, National University of Singapore, Singapore.}





\begin{abstract}
The neural network (NN)-based direct uncertainty quantification (UQ) methods have achieved the state of the art performance since the first inauguration, known as the lower-upper-bound estimation (LUBE) method. However, currently-available cost functions for uncertainty guided NN training are not always converging and all converged NNs are not generating optimized prediction intervals (PIs). Moreover, several groups have proposed different quality criteria for PIs. These raise a question about their relative effectiveness. Most of the existing cost functions of uncertainty guided NN training are not customizable and the convergence of training is uncertain. Therefore, in this paper, we propose a highly customizable smooth cost function for developing NNs to construct optimal PIs. The optimized average width of PIs, PI-failure distances and the PI coverage probability (PICP) are computed for the test dataset. The performance of the proposed method is examined for the wind power generation and the electricity demand data. Results show that the proposed method reduces variation in the quality of PIs, accelerates the training, and improves convergence probability from 99.2\% to 99.8\%.  
\end{abstract}



\end{frontmatter}

\begin{table*}[!t]   

\begin{framed}
\nomenclature{$ACE$}{Average coverage error}
\nomenclature{$ACE$}{Average coverage error}
\nomenclature{$CWC$}{Coverage width criterion}
\nomenclature{$CWFDC$}{Coverage width failure distance criterion}
\nomenclature{$DIC$}{Deviation information-based criterion}
\nomenclature{$LUBE$}{Lower upper bound estimation method}
\nomenclature{$MAPE$}{Mean absolute percentage error}
\nomenclature{$NN$}{Neural network}
\nomenclature{$PI$}{Prediction interval}
\nomenclature{$PIC$}{Prediction interval coverage}
\nomenclature{$PICP$}{Prediction interval coverage probability}
\nomenclature{$PINAFD$}{Prediction interval normalized average failure distance}
\nomenclature{$PINAW$}{Prediction interval normalized average width}
\nomenclature{$PINC$}{Prediction interval nominal coverage (1-$\alpha$)}
\nomenclature{$PIW$}{Prediction interval width}
\nomenclature{$RMSE$}{Root mean square error}
\nomenclature{$SSE$}{Sum squared error} 
\nomenclature{$UK$}{United Kingdom}
\nomenclature{$UQ$}{Uncertainty quantification}

\nomenclature{$||e||$}{Error quantity in L. G. Marn's cost function}
\nomenclature{$\epsilon_j$}{Difference between the target and the point prediction for $j^{th}$ sample}
\nomenclature{$\gamma^{(\alpha, PICP)}$}{The PI-failure penalty function}
\nomenclature{$\lambda ^{(\alpha)}$}{Variation to bound multiplier in traditional PIs}
\nomenclature{$\mu_j$}{Mean of the probability density for $j^{th}$ sample}  
\nomenclature{$\mu_{CWC}$}{Average CWC of converged NNs}
\nomenclature{$\mu_{CWFDC}$}{Average CWFDC of converged NNs}
\nomenclature{$\mu_{PICP}$}{Average PICP of converged NNs}
\nomenclature{$\mu_{PINAFD}$}{Average PINAFD of converged NNs}
\nomenclature{$\mu_{PINAW}$}{Average PINAW of converged NNs}
\nomenclature{$\sigma_j$}{Target variance}
\nomenclature{$\sigma_{\hat{y}_j}$}{Target variance for epistemic uncertainty}
\nomenclature{$\sigma_{\hat{\epsilon}_j}$}{Target variance for aleatory uncertainty}
\nomenclature{$\sigma_{PICP}$}{PICP variation}
\nomenclature{$c_j$}{The binary coverage value for $j^{th}$ sample}
\nomenclature{$CP()$}{Cumulative probability density function}
\nomenclature{$n$}{Number of samples}
\nomenclature{$\tilde{N}_{iter}^{(PICP(1\%))}$}{ \ \ Median of the number of iterations to reach $|1-\alpha + \delta -PICP| < 1\%$}
\nomenclature{$\tilde{N}_{iter}^{(PINAW(1.5))}$}{ \ \ \ \ Median of the number of iterations to reach $PINAW < 1.5 \times PINAW_{Opt}$}
\nomenclature{$N_L$}{The number of times $t_j < \underline{y}_j$}
\nomenclature{$N_{Trial}$}{Number of trial}
\nomenclature{$N_U$}{The number of times $t_j > \overline{y}_j$}
\nomenclature{$P_j()$}{Conditional probability function for $j^{th}$ sample}
\nomenclature{$PI_j^{(\alpha)}$}{PI of (1-$\alpha$) expected PICP for $j^{th}$ sample}
\nomenclature{$pun$}{Non-exponential penalty by G. Zhang}
\nomenclature{$PIW_j$}{PIW for $j^{th}$ sample}
\nomenclature{$R$}{Range of targets}
\nomenclature{$S_{AV}$}{The average interval score}
\nomenclature{$S_j$}{The interval score for $j^{th}$ sample}
\nomenclature{$t_j$}{Target for $j^{th}$ sample}
\nomenclature{$w$}{Weight in the NN}
\nomenclature{$y$}{Output in the sample}
\nomenclature{$\overline{y}_j$}{Upper bound for $j^{th}$ sample}
\nomenclature{$\underline{y}_j$}{Lower bound for $j^{th}$ sample}
\nomenclature{$\hat{y}$}{The true regression mean}

\nomenclature{$\alpha$}{Expected non-coverage probability}
\nomenclature{$\beta$}{PI coverage penalty factor}
\nomenclature{$\beta _1$, $\beta _2$}{Weighting factors in L. G. Marn's cost function}
\nomenclature{$\delta$}{Coverage margin}
\nomenclature{$\gamma, \lambda$}{Fitting parameters in C. Wan's cost function}
\nomenclature{$\eta$}{Hyperparameter for penalizing low coverages}
\nomenclature{$\rho$}{The failure distance resistance parameter}
\nomenclature{$\sigma_p$}{The penalty factor by G. Zhang}

\printnomenclature

\end{framed}

\end{table*}

\section{Introduction}
All-natural quantities have some uncertainties. The quantity may slightly or greatly vary for the same circumstances. The variance of the quantity at a circumstance is the level of uncertainty for the circumstance. The level of uncertainty is heteroscedastic. The predictability of the same quantity can be different based on circumstances. Traditional point prediction systems predict a value which is the most probable for the corresponding input combination. The actual value may differ from the prediction slightly or greatly based on circumstances \cite{kabir2018neural}. For instance, electricity demand at off-peak and full-peak hours is highly predictable but the demand on the transition times may largely vary from one day to another day \cite{ shrivastava2015prediction, koprinska2015correlation}.  The difference between the prediction and the actual value may be caused by the modeling error, or the inherent randomness of the system \cite{zhang2007statistical,  clements2004evaluating}. The inclusion of some inputs such as the current temperature and calendar information may reduce the uncertainty of predictions. However, some portion of the uncertainty can be random and cannot be predicted based on existing features.  The uncertainty can also be asymmetrically heteroscedastic and the point prediction with a certain error possibility cannot provide adequate information to the user \cite{kendall2017uncertainties, gal2016uncertainty}.

\begin{figure}
\begin{center}
\includegraphics[width=8.75cm]{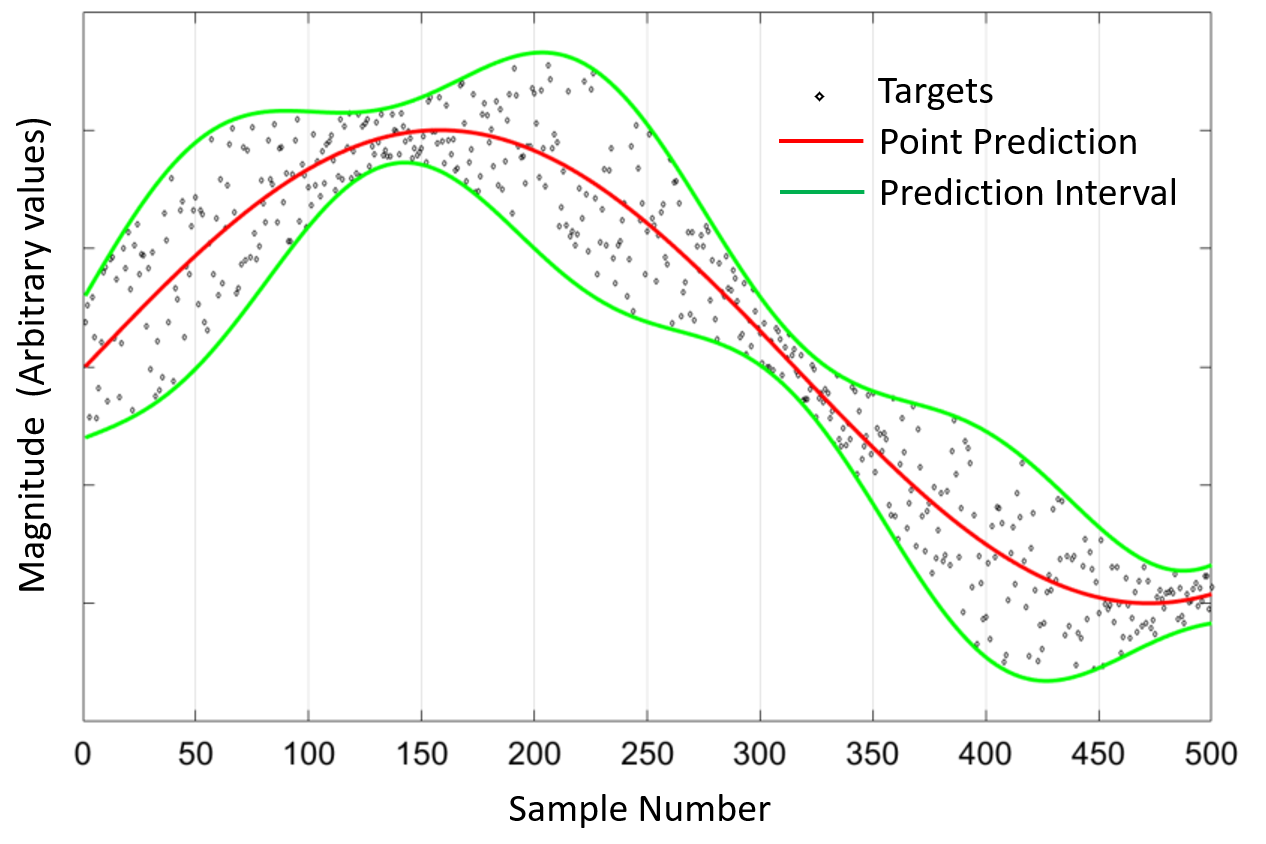}
\caption{\label{PI_Advantage} Importance of the uncertainty quantification. The point prediction, presented by the red line with a constant error possibility (the root mean square error) cannot represent the heteroscedastic uncertainty. PIs, represented by green lines becomes narrow in the less uncertain regions and become wide in more uncertain regions. Therefore, PIs can represent heteroscedastic uncertainty.}
\end{center}
\end{figure}

Probabilistic forecast, such as an uncertainty bound is also popular in decision-making \cite{roy2018new, laptev2017time}. However, a single uncertainty bound is unable to represent the level of uncertainty. Multiple uncertainty-bounds can be applied to quantify uncertainties. PI is a recognized UQ method, applies the upper and the lower bounds to quantify the level of uncertainty. Probabilistic forecasts such as prediction intervals (PIs) with a certain coverage probability are more appropriate for understanding the uncertain condition. Fig. \ref{PI_Advantage} presents the uncertainty captured by PIs with 99\% coverage probability. The probability density function changes from sample to sample \cite{quan2014uncertainty1, quan2016integration}. The width of the PI varies from sample to sample based on the corresponding uncertainty.  Decision-makers get the most probable regions of targets from PIs, generated by NNs even for an asymmetric and heteroscedastic system \cite{ ak2018adequacy}.

NN is commonly recognized as a black box to its end-user. It can provide state-of-art performance with proper training. The designer of any NN chooses NN-size, activation function and initial weights. Initial weights can also be random or pre-defined. Proper training provides an optimal selection of weights on interconnects and biases. The optimal NN output is a weighted sum of inputs and functions of inputs. Weight optimization is performed through a reward-based system. The reward is calculated based on the performance of NN through a cost function in each cycle. Therefore, a cost function needs to be designed considering the purpose, quality criteria, critical situations, and the convergence of optimization \cite{altinten2008self}.

This paper proposes an optimal PI construction technique considering different aspects of recently proposed NN-based direct PI construction techniques. The direct construction of PIs from the NN result in a sharp PI for any type of probability distributions, such as skewed-Gaussian, log-normal and multimodal. Several direct PI construction techniques have been proposed in the literature. The relative performance of these techniques is questionable and should be comprehensively investigated. Every new algorithm is claimed to be the best with data analysis during its proposal. However, the result may vary with different datasets. Therefore, we discuss the philosophy of developing cost functions and provide a novel one. The proposed cost function combines the important philosophies of recently proposed NN-based direct PI computation methods. 
The paper presents a rigorous performance analysis for the wind power generation and electricity demand data. The method is also applicable for the UQ of several other datasets, such as electricity prices, and other renewable generations. The improvements in the convergence for electricity demand, hydro and solar generations are also analyzed. Moreover, weather, geographical positions, and various human-made events can be considered as an input of the NN. The effect of many events may not be expressed through mathematical equations. NNs can find all of those hidden relations with reward-based training.

The paper is organized with the following flow of information. Section 2 presents the basics of the uncertainty quantification and the advantage of the NN-based direct uncertainty quantification. Section 3 presents all NN-based direct PI construction methods.  Section 4 presents the proposed PI construction method. Section 5 reports the simulation results and performance metrics. Section 6 is the concluding section.

\section{Uncertainty and its Quantification}
\subsection{Increased Uncertainty in Power Grid}
All real-world events consist of sub-events, among which many are random. However, their combined effect can be interval predictable or even deterministic. When tossing a coin for a single time, the probability of getting the head or the tail is equal. However, the outcome of one hundred tossings of a coin is quite predictable. There is a 97.9\% chance of getting 40 to 60 heads and there is 72.88\% chance of getting 45 to 55 heads. Therefore, 20\% region of the output range contains about 97-98\% of the probability density. Therefore, that outcome is interval predictable. The outcome of ten thousand tossings is more predictable. There is a 95.56\% chance of getting 4900 to 5100 heads. Therefore, 2\% region of the output range contains about 95-96\% of the probability density and it is point predictable with 1-2\% root mean squared error (RMSE) \cite{kabir2019partial}.  

A grid is connected to millions of electrical appliances. Whether individual equipment is consuming electricity or not is difficult to predict. However, the total electricity consumption of a large grid is point predictable. A fair coin has an equal probability of head or tail but an appliance may have a 10\% probability of consuming electricity and that probability varies depending on time or weather or any other conditions. Therefore, the total power consumption of a large grid is predictable and depends on major events, such as weather, time, vacation, sports, etc.  

Let us consider two events: the percentage of heads in tossing 1) five coins, 2) ten thousand coins. When the first event has a higher weight than the second event on a quantity, the quantity becomes highly uncertain. The grid also contains elements of different predictability. Wind power generation is highly random. Installation of large scale wind power plants has made the overall generation more unpredictable. The large-scale introduction of the electric vehicle has made the overall consumption more unpredictable.

\subsection{Uncertainty Quantification}
All systems have some inherent randomness, known as the aleatory uncertainty. The output value slightly or greatly varies for the same input combination. The other type of uncertainty is epistemic or subjective which can be properly captured by the precise modeling \cite{ayyub2010uncertainty, alam2016effect}. This uncertainty can cause significant prediction error when several secondary or tertiary effects are overlooked during the modeling. The future value of the parameter or the target ($t_j$) is represented as:
\begin{equation}
        t_j=y_j+\epsilon_j
        \label{eq:regressionEq1}
\end{equation}
where $\epsilon_j$ is the zero expectation error signal and $j(=1,2,\cdots n)$ is the sample number. Therefore, total uncertainty is represented as follows:
\begin{equation}
		t_j-\hat{y}_j=[y_j-\hat{y}_j]+\epsilon_j
		\label{eq:regressionEq2}
\end{equation}
where $\hat{y}_j$ is the true regression mean for the $j^{th}$ sample. 
When the two terms in (\ref{eq:regressionEq2}) are independent, the total variation associated with the model outcome is represented as:
\begin{equation}
        \sigma_j^2=\sigma_{\hat{y}_j}^2+\sigma_{\hat{\epsilon}_j}^2
        \label{eq:regressionEq3}
\end{equation}
The term $\sigma_{\hat{y}_j}^2$ represents the subjective uncertainty and $\sigma_{\hat{\epsilon}_j}^2$ represents the inherent randomness.

Point prediction is the most widely used approach for predicting unknown quantities. It provides a value corresponding to the lowest statistical error. Statistical errors are measured by several criteria, such as the RMSE, mean absolute percentage error (MAPE) or sum squared error (SSE). Therefore, the user lacks information about the level of heteroscedastic uncertainty ($\sigma^2_j$). An interval forecast with a certain probability of coverage expresses both uncertainties. 

The PI for $j^{th}$ sample is traditionally represented as follows \cite{johnson2001introduction}:
\begin{equation}\label{Eq:PI_trad}
        PI_{j, \ Traditional}^{(\alpha)} = [\mu_j-\lambda ^{(\alpha)} \sigma_j,  \ \  \mu_j+\lambda ^{(\alpha)} \sigma_j]
\end{equation}
Here, $\mu_j$ is the mean and $\lambda ^{(\alpha)}$ is the target coverage probability parameter. The value of $\lambda ^{(\alpha)}$ is 1.15, 1.64 and 1.96 respectively for 75\%, 90\% and 95\% theoretical coverage \cite{kirkwood2010essential}.
Therefore, the width of the PI of $j^{th}$ sample is as follows:
\begin{equation}
        PIW_j = 2 \lambda ^{(\alpha)} \sigma_j 
\end{equation}

The distribution of uncertainties may not be purely Gaussian for any combination of inputs. Some distributions can be log-normal, skewed Gaussian, or multimodal. Therefore, PIs constructed through the Gaussian assumption fails to maintain a narrow width and the required coverage simultaneously. 

A relatively smarter approach to constructing PIs is considering the cumulative distribution function.
Conditional probability functions can construct an interval of ($1-\alpha$) confidence level.
The upper bound ($\overline{y}_j$) is a value greater than  ($1-\alpha/2$) portion of the probability density function. Therefore, $\overline{y}_j$ can be represented as \cite{pinson2010conditional}:
\begin{equation}
		P_j(t_j<\overline{y}_j) = 1-\alpha/2
		\label{eq:CumPEq1}
\end{equation}
where $P_j(condition)$ is the probability function for the target at $j^{th}$ sample ($t_j$). The cumulative probability density function ($CP$) can represent the relation as follows:
\begin{equation}
		CP(\overline{y}_j) = 1-\alpha/2
		\label{eq:CumPEq2}
\end{equation}
Taking inverse:
\begin{equation}
		\overline{y}_j = CP^{-1}(1-\alpha/2)
		\label{eq:CumPEq3}
\end{equation}
The lower bound ($\underline{y}_j$) and the upper bound ($\overline{y}_j$) form the PI. Therefore, the PI (=$[\underline{y}_j, \overline{y}_j]$) becomes as follows:
\begin{equation}
		PI_{j, \ Conditional}^{(\alpha)} = [CP^{-1}(\alpha/2), CP^{-1}(1-\alpha/2)]
		\label{eq:CumPEq4}
\end{equation}

The direct NN-based PI computation techniques construct an optimal PI empirically without considering any theoretical probability distribution. 

\begin{figure}
\begin{center}
\includegraphics[width=7.5cm]{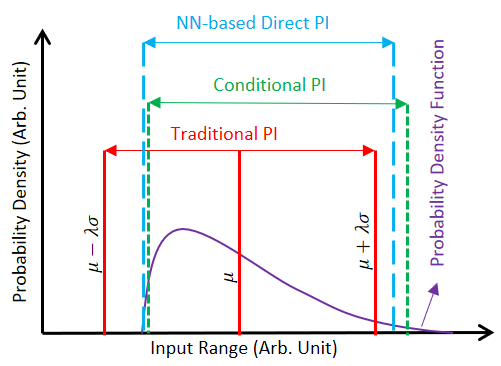}
\caption{\label{PDF_to_PI}The advantage of the NN-based direct PI construction over traditional and conditional PI construction techniques for a non-Gaussian probability distribution. The NN-optimization technique finds an optimal PI for any arbitrary probability distribution. }
\end{center}
\end{figure}
Fig. \ref{PDF_to_PI} presents the advantage of the NN-based direct PI construction over traditional and conditional PI construction techniques for a non-Gaussian probability distribution, known as asymmetric heteroscedastic uncertainties \cite{carnero2016identification}. Although, there exist multiple variables in our NN based PI construction system, Fig. \ref{PDF_to_PI} presents a one-dimensional input range for an easier visualization \cite{kabir2019partial}. Traditional and conditional PIs follow Eq. (\ref{Eq:PI_trad}) and Eq. (\ref{eq:CumPEq4}) respectively and the LUBE NN is directly trained to optimize the cost function. Conditional PI is optimal compared to traditional PIs. The NN-based direct PI moves upper and lower bounds slightly and evaluates the effect during the training to achieve an optimal PI coverage probability (PICP).

\section{Direct NN-based PI Construction Methods}
Several direct NN-based methods have been proposed by different groups. The current section discusses relevant methods where a cost function is proposed or modified to overcome a limitation. 

\subsection{Initial LUBE Method}
The lower upper bound estimation ($LUBE$) method is the very first method for constructing PIs through direct training \cite{khosravi2010construction}. The LUBE method is based on the following philosophies:
\begin{itemize}
    \item The PI should cover samples with an equal or higher probability compared to the PI nominal coverage (PINC = $1-\alpha$).
   \item When the required PI coverage probability (PICP) is achieved (PICP $\geq$ PINC), the quality of PI depends on the narrowness of the average width.  
\end{itemize}

The LUBE method minimizes the coverage width criterion (CWC). PICP, PINAW, and CWC are defined as follows:
\begin{equation}
\label{eq:PICP}
PICP = \frac{1}{n} \sum_{j=1}^n c_j
\end{equation}
provided that,
\begin{equation}
c_j = 	\begin{cases}
		1, t_j \in [\underline{y}_j, \overline{y}_j] \\
		0, t_j \not \in [\underline{y}_j, \overline{y}_j].
		\end{cases}
\end{equation}
\begin{equation}
\label{Eq:PINAW}
PINAW = \frac{1}{n R} \sum_{j=1}^n (\underline{y}_j-\overline{y}_j).
\end{equation}
\begin{equation}\label{Eq:CWC}
CWC = PINAW \{1 + \gamma^{(\alpha, PICP)} e ^{\eta (PINC - PICP)}\}
\end{equation}
provided that,
\begin{equation*}
\gamma^{(\alpha, PICP)} = 	\begin{cases}
		1, PICP < PINC \\
		0, PICP \geq PINC
		\end{cases}
\end{equation*}
$\eta$ =50 is a hyperparameter used for penalizing low coverages. 
The LUBE method also proposes the structure of the NN. Fig. \ref{Optimal_PI_NN} presents the structure of a lower-upper-bound-based NN. In the LUBE NN, input and hidden layers are shared. Therefore, the difference between the lower and upper bounds is the weights of the connections between the last hidden layer to the output layer and output layer biases.

\subsection{$LUBE$ with Independent Width and Penalty Factors}
The width factor (PINAW) exists in a multiplicative manner with the penalty factor ($\gamma^{(\alpha, PICP)}$), as shown in (\ref{Eq:CWC}) in the initial LUBE method. Therefore, the optimization was controversial \cite{pinson2014discussion, wan2014discussion}. One of the concerns is the optimization at zero width (PINAW = 0), that happens frequently with the initial LUBE method. Khosravi et. al. resolved most of those controversies with the proposal of a new CWC equation of independent width and penalty factor \cite{khosravi2014closure} as presented at Eq. (\ref{Eq:CWC2}). 
\begin{equation}\label{Eq:CWC2}
CWC = PINAW  + \gamma^{(\alpha, PICP)} e ^{\eta (PINC - PICP)}
\end{equation}
The function is modified based on the following philosophy:
\begin{itemize}
    \item A separate PINAW in an additive manner restricts optimization at PINAW = 0.
\end{itemize}

\subsection{$LUBE$ with Continuous Cost Function}
The NN training in LUBE method often fails to converge with the traditional LUBE cost function. Therefore, the following equation is recently introduced to achieve a continuous cost function at (PICP = PINC) \cite{kabir2018adversarial}:
\begin{equation}\label{Eq:d2}
CWC = PINAW + \gamma^{(\alpha, PICP)} 
\end{equation}
provided that,
\begin{equation*}
\gamma^{(\alpha, PICP)} = 	\begin{cases}
		e ^{\eta (PINC - PICP)} - 1, PICP < PINC \\
		0, PICP \geq PINC
		\end{cases}
\end{equation*}
where $\gamma^{(\alpha, PICP)}$ is the PI-failure penalty function. The value of that function is called the penalty. $\gamma^{(\alpha, PICP)}$ is zero for (PICP $\geq$ PINC). In contrast, its value exponentially increases with the lowering of PICP when  PICP $<$ PINC.

The concept of a continuous cost function is well known in the model-design rules of circuit simulations \cite{coram2004and, kabir2016modeling1}. The philosophy behind the continuous cost function is as follows:
\begin{itemize}
   \item A continuous model is the prerequisite for the convergence of the simulation. Especially when the model is applied to optimize anything through an iterative process.
   \item A discontinuity on the model or it's derivative may result in a very large gradient ($\partial y/ \partial w$) during an iteration, as the $\Delta y$ remains large for a very small $\Delta w$ at the point of discontinuity. Therefore, the discontinuity may potentially result in the overflow while nearby regions are mapped following the gradient. 
\end{itemize}
As the most optimized point of the NN training is PICP = PINC, the step-sizes are reduced near that point to achieve optimized weight with high precision \cite{magoulas1997effective, magoulas1999improving}. That means, $\Delta x$ or a component of $\Delta x$ may become very small while iterating near the most optimized point. Therefore, the cost function needs to be smooth near the minima.

\subsection{Can Wan's Interval Score-based Cost Function}
C Wan et al. \cite{kabir2018neural} also propose a slightly different cost function for the NN-based direct interval forecasting. They consider both width dependent and coverage dependent components. The average coverage error (ACE) is the coverage dependent component in C Wan's cost function. ACE is defined as follows:
\begin{equation}\label{Eq:ACE}
ACE = PICP - PINC
\end{equation}
where $PINC = 1-\alpha$ is the target PI coverage, known as the PI nominal coverage. PI-width of the $j^{th}$ sample is defined as follows:
\begin{equation}\label{Eq:PIW_J}
PIW_j = \overline{y}_j-\underline{y}_j
\end{equation}
C Wan et al. \cite{wan2013direct} also define a component named the interval score. The interval score for $j^{th}$ sample is as follows:
\begin{equation}
S_j = 	\begin{cases}
		-2 \alpha \times PIW_j -4(\underline{y}_j - t_j), t_j< \underline{y}_j \\
		-2 \alpha \times PIW_j,  \overline{y} \geq t_j \geq \underline{y}_j\\
		-2 \alpha \times PIW_j -4(t_j - \overline{y}_j), t_j > \overline{y}_j 
		\end{cases}
\end{equation}
The average interval score is as follows:
\begin{equation}
\label{eq:S_AV}
S_{AV} = \frac{1}{n} \sum_{j=1}^n S_j
\end{equation}
With the weighted summation of ACE and the interval score, the C Wan's cost function is as follows:
\begin{equation}
\underset{NN}{\text{min}} \ \ \lambda |S_{AV}| + \gamma |ACE|.
\end{equation}
Both $\lambda$ and $\gamma$ are set to one to provide the equal weight towards the PICP calibration and the PI sharpness.

According to our analysis, that cost function is based on the following philosophies:
\begin{itemize}
    \item Both high and low PICPs are penalized with equal concentration to maintain a gradient throughout the input domain.
   \item Both of the width of the interval ($PIW_j$) and the failure distance ($\underline{y}_j - t_j$ or $t_j - \overline{y}_j$) are optimized.
   \item The failure distance is considered with much higher priority ($2/\alpha$ times) compared to the width.
\end{itemize}


\subsection{L. G. Mar´ıns' Deviation from Mid Interval Consideration}
L. G. Mar´ın et al. \cite{marin2016prediction} consider the deviation from the mid interval and a continuous cost function. Their cost function is as follows:
\begin{equation}
\underset{NN}{\text{min}} \ \ \beta _1 PINAW + \beta_2 ||e||^2 + exp[-\eta \big(PICP-(1-\alpha)\big)]
\end{equation}
They also perform the normalization of $||e||^2$ and PINAW. They introduce the deviation from mid interval as $||e||$ and named it as the error quantity defined as follows: 
\begin{equation}
||e|| = \sqrt{ \sum_{j=1}^n \Big|t_j- \frac{\overline{y}_j + \underline{y}_j}{2}\Big|^2 }.
\end{equation}
The function is proposed based on the following philosophy:
\begin{itemize}
    \item The most probable region of the target may not stay near the middle of the interval in the direct PI construction method. Minimization of the deviation of the target from the center can potentially shift the most probable region near the center of PIs. 
\end{itemize}


\subsection{G. Zhang's Deviation Information-based Criterion (DIC)} 
\begin{figure}
\begin{center}
\includegraphics[width=7.5cm]{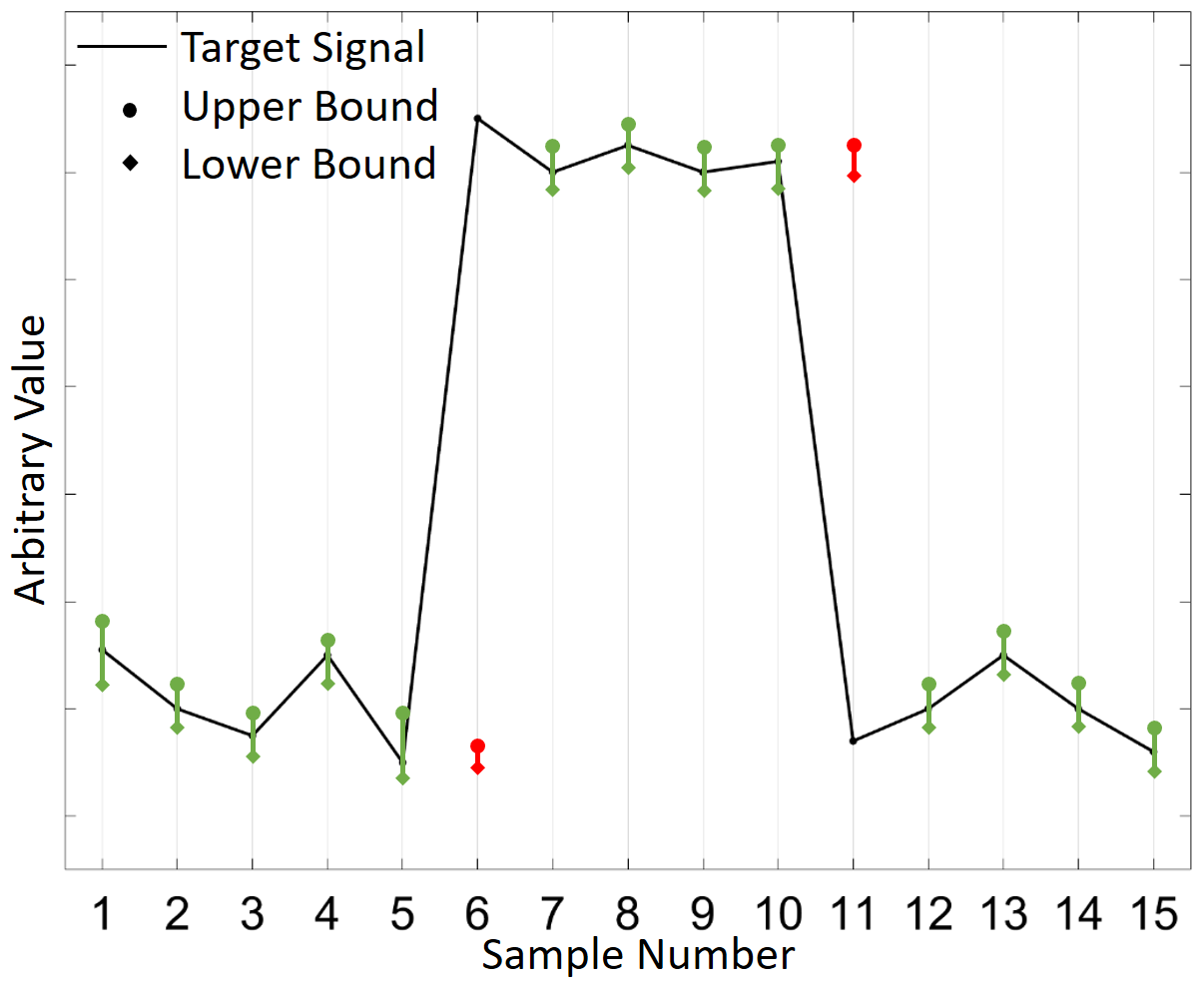}
\caption{\label{CWC_lim} A rough diagram presenting NN based PIs when the optimization considers only  PICP and PINAW. PIs cover 80\% to 90\% targets but fail to predict sharp changes. Successful and unsuccessful PIs are represented by green and red lines respectively with upper and lower bound marks.}
\end{center}
\end{figure}

When the cost function only considers PICP and PINAW, the NN finds smart ways to optimize the cost function. Though the PINAW becomes much smaller, the NN often keeps a smaller width instead of covering the target in critical situations. Therefore, target PICP is achieved with much smaller PINAW. However, PI misses the target by a large distance in critical situations. The situation is illustrated with a rough drawing in Fig. \ref{CWC_lim}. Intervals covering and missing the target are denoted by the green color and red color respectively. In the critical situation, the NN aims to narrow down the PI instead of covering it by increasing the width. However, the user of the prediction algorithm needs indications of sudden rises or falls with higher accuracy to manage critical situations.

G. Zhang et al. \cite{zhang2015advanced} tried to avoid the computation extensive exponential cost function. They proposed deviation information-based criterion defined as:
\begin{equation}\label{Eq:DIC}
DIC = PINAW + \gamma^{(\alpha, PICP)}. pun
\end{equation}
provided that,
\begin{equation*}
\label{eq:pun}
pun = \sigma_p \sum_{j=1}^{N_L} (\underline{y}_j-t_j) + \sigma_p \sum_{j=1}^{N_U} (t_j-\overline{y}_j)
\end{equation*}
and
\begin{equation*}
\gamma^{(\alpha, PICP)} = 	\begin{cases}
		1, PICP < PINC \\
		0, PICP \geq PINC
		\end{cases}
\end{equation*}
The cost function is modified based on the following philosophy:
\begin{itemize}
    \item Avoiding computationally expensive exponential function.
    \item Bringing the most probable region of the target near the middle of the PI.
\end{itemize}



\section{Proposed Method}
In this paper, a smooth and customizable cost function is proposed for the uncertainty guided NN training. Uncertainty guided NNs predict the upper bound and the lower bound. Bounds are computed with NNs without any assumption on the distribution.

Studying motives of all proposed algorithms, the following key criteria of a good cost function are concluded:

\begin{enumerate}
  \item PICP, PINAW and PI normalized average failure distance (PINAFD) are important parameters of an ideal cost function. The consideration of the deviation from the mid-interval also results in a slightly lower failure distance but the PI becomes much wider.
  \item Cost function and its derivatives need to be continuous for a better convergence profile of the NN-training.    
  \item Often a trained NN fails to maintain $PICP \geq PINC$ with the test dataset due to the slight variation between the training and test datasets. Therefore, a small coverage margin ($\delta$) is required during the training.
  \item The entire input domain needs to have single minima at $PICP = PINC + \delta$. Except at the minima, the input domain needs to have non-zero gradients directing the optimization towards the minima.
  \item A simpler or less computation extensive cost function is preferred. However, the simplicity of the function is not related to the quality of PIs.
  \item Different users may have different preferences towards width, failure distance, and coverage penalty. 
\end{enumerate}

PIs may fail to cover the target during a sudden fluctuation but the NN should try to bring the nearby bound of the PI close to the target. When the nearby bound of the PI is close to the uncovered target, the user can manage the situation with a lower difficulty. The proposed NN training considers the minimization of the prediction interval normalized failure distance (PINAFD). The expression of PINAFD is as follows:
\begin{equation}
\label{Eq:PINAFD}
PINAFD = \frac{\sum_{j=1}^n (1-c_j) \times min(|t_j-\overline{y}_j|,|\underline{y}_j-t_j|)}{R \times \Big\{\sum_{j=1}^n (1-c_j)\Big\} +\epsilon } .
\end{equation}
where $c_j$ contains the same meaning as of Eq. (\ref{eq:PICP}). The minimum distance of the target from bounds is considered to be the failure distance for the corresponding sample when the target is not bounded. The total failure distance is divided by the total number of missing samples ($\sum_{j=1}^n (1-c_j)$) and normalized by the range ($R$) to achieve normalized average value. $\epsilon$ is a small value to avoid an undefined PINAFD value for 100\% PICP. During an iteration of the training, NN may cover all samples, resulting in $\sum_{j=1}^n (1-c_j) = 0$. In such a situation, the value of PINAFD becomes zero by zero.  To avoid that undefined value, a small value, $\epsilon = 1e-10$ is added to the denominator to make zero PINAFD for 100\% PICP.

We formulate the proposed optimization parameter by adding a weighted PINAFD. The proposed optimization parameter is as follows:
\begin{equation}\label{Eq:CWFDC}
\underset{NN}{\text{min}} \ \ PINAW +\rho . PINAFD + \beta \times (1-\alpha+\delta -PICP)^2
\end{equation}
where $\rho$ is the failure distance resistance parameter, $\beta$ is the PI coverage penalty factor, and $\delta$ is the coverage margin. Usually, $\rho$ is set to one to provide an equal concentration towards the width and the failure distance. The coverage penalty $\beta$ needs to be high enough to provide higher concentration towards PICP than a slight shrink of average width or failure distance. Therefore, the value of $\beta$ is usually set to more than 200. The test PICP can be slightly lower than the train PICP due to a slight variation among datasets. The initial LUBE method strictly maintains the PICP higher than the nominal PICP. However, in a linear or a polynomial penalty, the test PICP can be slightly lower than the nominal. Therefore, a slight margin of $\delta = \alpha/50$  is kept. The variation of PICP is related to the nominal PICP. When $\alpha$ is small, the sample density near the edge of PIs is lower and the variation in PICP is also lower. With $\alpha = 5\%$, a slight margin of $\delta = 0.1\%$ is enough for obtaining a test PICP of 95\% most of the time. Therefore, $\delta = \alpha/50$ is kept for increasing the margin with increasing $\alpha$.

The proposed optimization parameter, presented in Eq. (\ref{Eq:CWFDC}) is named as the Coverage Width Failure Distance Criteria (CWFDC). The proposed cost function is considering the coverage (PICP), the normalized average width (PINAW),  and the failure distance (FD). Besides these considerations, it is smooth and considers a small margin to withstand a slight PICP variation.

\section{Result Evaluation}
\begin{figure}
\begin{center}
\includegraphics[width=7.7cm]{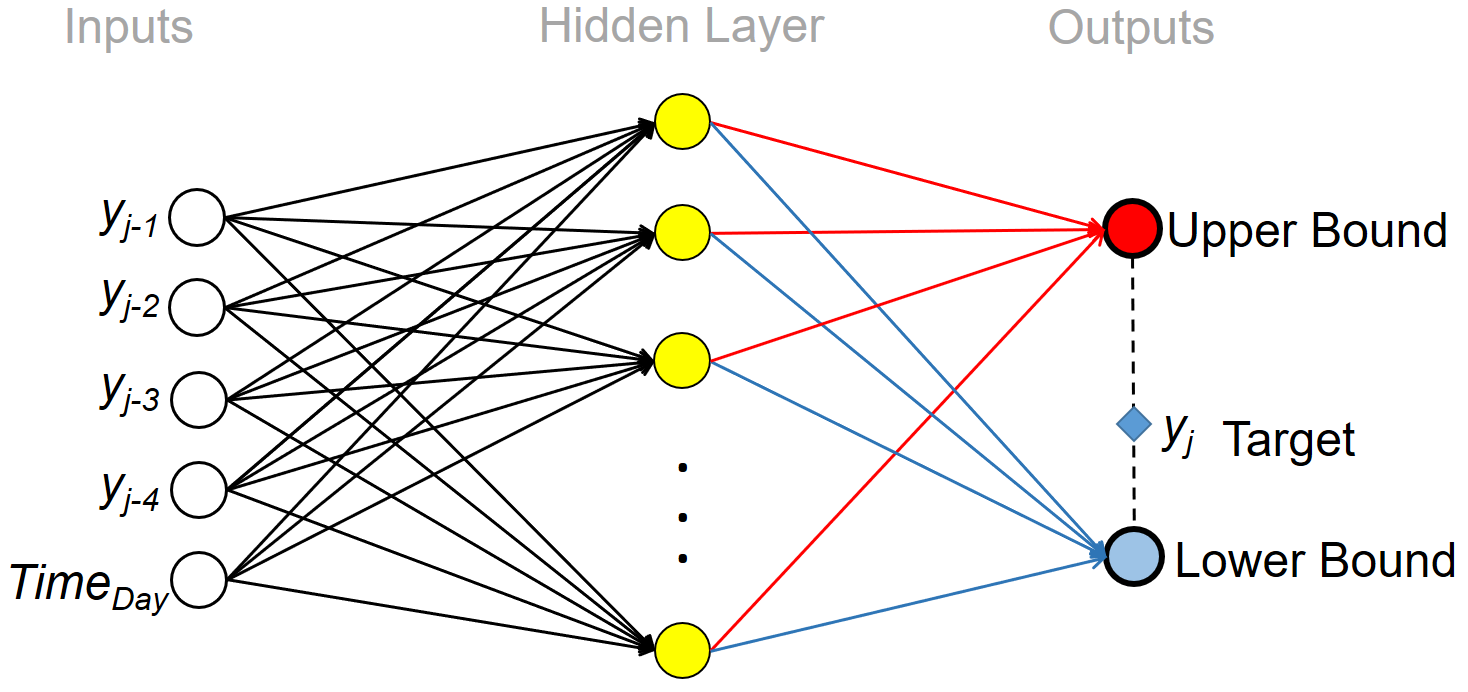}
\caption{\label{Optimal_PI_NN}The structure of the NN with input-output combinations. Four recent samples and the time is applied to quantify the uncertainty on the next sample.}
\end{center}
\end{figure}

The wind power generation and the electricity demand samples from August 2012 to August 2019 are downloaded from the UK-grid website \cite{UK_GW}. Four recent samples and the time in the hour ($Time_{Day} = hour + minutes/60$) on a corresponding day is provided as the input to the NN. Fig. \ref{Optimal_PI_NN} presents the structure of the NN with input-output combinations. We apply the simulated annealing technique for NN training. NNs of different sizes and initializations are trained with different cost functions to evaluate the result. Different steps of the result evaluation are as follows:  
\begin{enumerate}
    \item Finding the optimal NN-sizes for different cost functions, different PICP, and different datasets.
    \item Evaluation of PINAW, PICP, and PINAFD for NNs with optimal sizes.
\end{enumerate}

The NN size optimization is vital to avoid overfitting or underfitting \cite{blei2003latent}. The optimal neuron size of any NN-based prediction system depends on both the data and the cost function. Therefore, size of NNs is optimized for each cost function at first. Then, NNs of optimized sizes are trained to construct PIs and to evaluate its performance for the test set. 

\subsection{The LUBE method}
\begin{figure}
\begin{center}
\includegraphics[width=8.75cm]{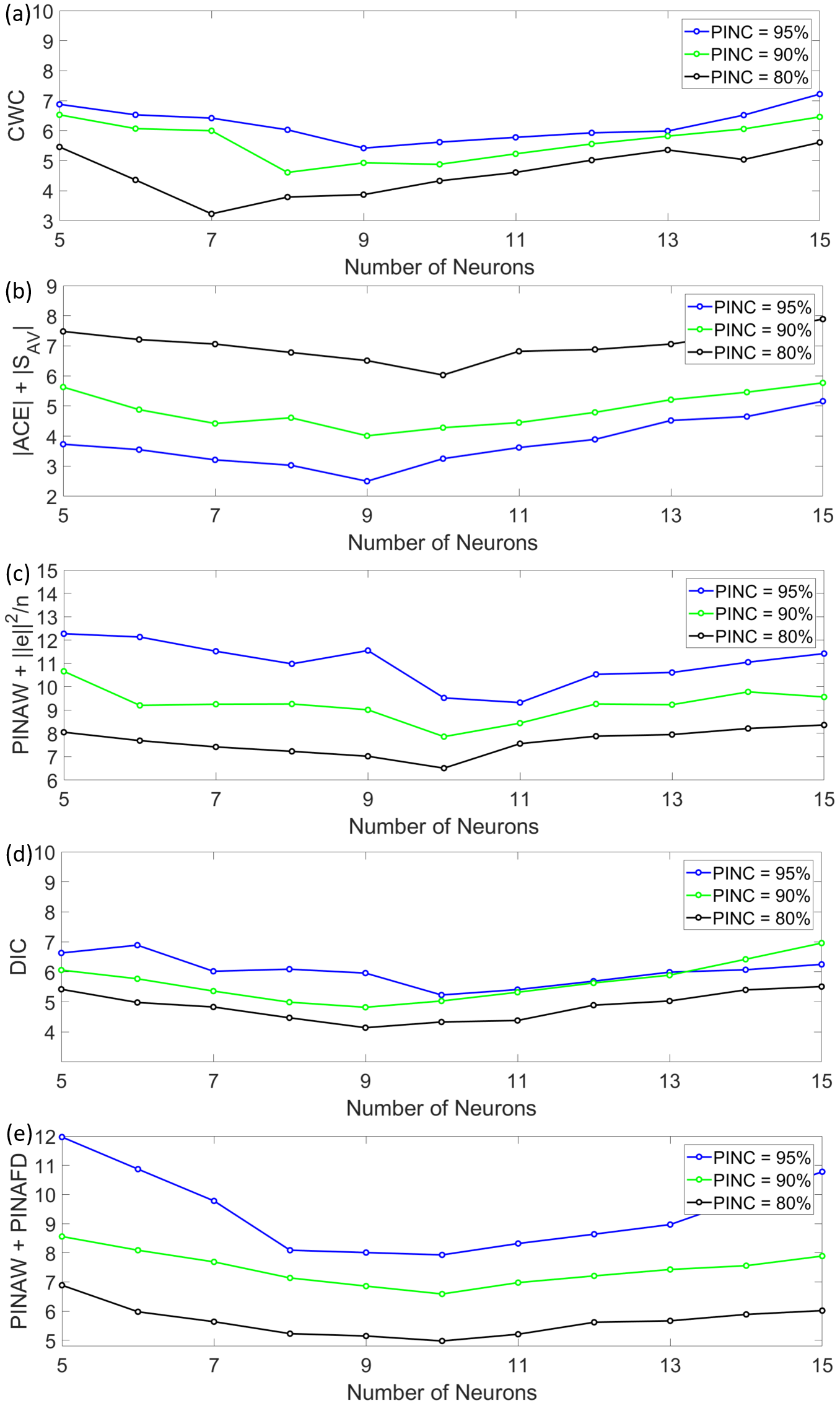}
\caption{\label{Optimal_c}NN-size optimization for the uncertainty quantification of the wind power generation of the UK grid.  (a) The LUBE cost function. Optimized NN-sizes are 9 for $\alpha$=5\%, 8 for $\alpha$=10\% and 7 for $\alpha$=20\%. (b) C. Wan's cost function. Optimized NN-sizes are 10 for $\alpha$=5\%, 9 for $\alpha$=10\% and 9 for $\alpha$=20\%. (c) L. G. Marn's cost function. Optimized NN-sizes are 11 for $\alpha$=5\%, 10 for $\alpha$=10\% and 10 for $\alpha$=20\%. (d) G Zhang's cost function. Optimized NN-sizes are 10 for $\alpha$=5\%, 9 for $\alpha$=10\% and 9 for $\alpha$=20\%. (e) the proposed cost function. Optimized NN-size is 10 for $\alpha$=5\%, 10\%, and 20\%.}
\end{center}
\end{figure}

Single hidden layer NNs with different neuron numbers are trained to find an optimal NN size. The NN is trained with four random initializations and one with the lowest CWC is selected. Fig. \ref{Optimal_c}(a) presents the lowest CWC values for different NN sizes and three different PINC values for wind power generation data. The number of neurons is varied between 5 and 15. The optimal NN size is found to be 9, 8, and 7 for PINC = 95\%, 90\%, and 80\% respectively. Similarly, the optimum size of NNs is found to be 8, 8 and 6 for  PINC = 95\%, 90\%, and 80\% respectively for the electricity demand data.

Khosravi et al. \cite{khosravi2014closure} previously observed that the training of the LUBE method fails to converge once in twelve cases. Therefore, the training converges at roughly 91.7\% situations. 5.2\% converged training provide a much wider or too narrow PIs. Therefore, only acceptable PIs are considered. The performance of the LUBE method is presented as the first segment of Table \ref{TABComp1}. Five hundred NNs are trained for each of the wind power and the electricity demand data and NNs providing logical PIs are considered. The LUBE method generates high-quality PIs on average. However, some PIs exhibit correct PICP on cross-validation data but provide slightly lower PICP compared to PINC for the test data. This happens due to the slight variation between the test dataset and the cross-validation dataset.

\begin{table*}
		\caption{PI Optimization Performance for the 5-minutes ahead forecast on the Wind Power Generation and Electricity Demand Data of the UK grid.}
		\label{TABComp1}
		\centering
			\begin{tabular}{|c|c|c|c|c|c|c|c|c|c|c|}	\hline
				
			\multicolumn{11}{|c|}{$LUBE$ Method}\\ \hline
				Data &$1-\alpha$    &$N_{Trial}$    &$\mu_{PINAW}$    &	$\mu_{PICP}$ 	& $\sigma_{PICP}$   & $\mu_{PINAFD}$ & $\mu_{CWC}$ & $\mu_{CWFDC}$ & $\tilde{N}_{iter}^{(PICP(1\%))}$ & $\tilde{N}_{iter}^{(PINAW(1.5))}$\\ \hline
				\multirow{ 3}{*}{\rotatebox[origin=c]{90}{\parbox[c]{1cm}{\centering Wind Power}}}
				&0.95&500	    &5.42		&	95.14   & 1.42    & 1.23 &5.42 &8.25 &91 & 126   \\ \cline{2-11}
				&0.90&500	    &4.61		&	90.12   & 2.61    & 2.08 &4.62 &13.09 &113 & 131  \\ \cline{2-11}	
				&0.80&500	    &3.23		&	80.15   & 2.32    & 2.74 &3.24 &28.47 &119 &137	 \\ \hline 
				
				\multirow{ 3}{*}{\rotatebox[origin=c]{90}{\parbox[c]{1cm}{\centering E. De-mand}}}
				&0.95&500	    &2.07	&	95.06& 0.70& 0.51& 2.09& 3.11 &78 & 104  \\ \cline{2-11}
				&0.90&500	    &2.01	&	90.09& 1.26& 0.74& 2.05& 4.08 &92 & 119 \\ \cline{2-11}	
				&0.80&500	    &1.90	&	80.14& 1.53& 0.77& 1.96& 8.27 &99 & 127 \\ \hline 
				\hline
				
				\multicolumn{11}{|c|}{Direct Interval Forecasting by Can Wan}\\ \hline
				Data &$1-\alpha$    &$N_{Trial}$    &$\mu_{PINAW}$    &	$\mu_{PICP}$ 	& $\sigma_{PICP}$   & $\mu_{PINAFD}$ & $\mu_{CWC}$ & $\mu_{CWFDC}$ & $\tilde{N}_{iter}^{(PICP(1\%))}$ & $\tilde{N}_{iter}^{(PINAW(1.5))}$\\ \hline
				\multirow{ 3}{*}{\rotatebox[origin=c]{90}{\parbox[c]{1cm}{\centering Wind Power}}}
				&0.95&500	    &7.74		&	95.11   & 0.33    & 0.41 &7.74 &8.25 &47 & 106  \\ \cline{2-11}
				&0.90&500	    &6.87		&	90.31   & 0.52    & 0.39 &6.87 &19.36 &51 & 105   \\ \cline{2-11}	
				&0.80&500	    &5.06		&	80.24   & 0.97    & 0.24 &5.06 &30.9 &56 & 110 	 \\ \hline 
				\multirow{ 3}{*}{\rotatebox[origin=c]{90}{\parbox[c]{1cm}{\centering E. De-mand}}}
				&0.95&500	    &2.59 &95.09& 0.31& 0.29&2.51 &3.19 &41 & 88   \\ \cline{2-11}
				&0.90&500	    &2.41 &90.04& 0.44& 0.22&2.44 &3.64 &46 & 97  \\ \cline{2-11}	
				&0.80&500	    &2.32 &80.01& 0.61& 0.18&2.36 &7.36 &50 & 102	 \\ \hline 
				\hline
				\multicolumn{11}{|c|}{Mid-interval Deviation Consideration by L. G. Mar´ın}\\ \hline
				Data &$1-\alpha$    &$N_{Trial}$    &$\mu_{PINAW}$    &	$\mu_{PICP}$ 	& $\sigma_{PICP}$   & $\mu_{PINAFD}$ & $\mu_{CWC}$ & $\mu_{CWFDC}$ & $\tilde{N}_{iter}^{(PICP(1\%))}$ & $\tilde{N}_{iter}^{(PINAW(1.5))}$\\ \hline
				\multirow{ 3}{*}{\rotatebox[origin=c]{90}{\parbox[c]{1cm}{\centering Wind Power}}}
				&0.95&500	    &9.02		&	95.21   & 1.42    & 1.09 &9.02 &9.21 &68 & 98   \\ \cline{2-11}
				&0.90&500	    &7.36		&	90.22   & 1.51    & 1.56 &7.36 &9.32  &64 & 103  \\ \cline{2-11}	
				&0.80&500	    &6.51		&	80.30   & 2.03    & 1.83 &6.51 &18.34 &64 & 107 	 \\ \hline 
				\multirow{ 3}{*}{\rotatebox[origin=c]{90}{\parbox[c]{1cm}{\centering E. De-mand}}}
				&0.95&500	    &3.03 &	95.16& 0.72& 0.39&2.89 &3.33 &42 &81   \\ \cline{2-11}
				&0.90&500	    &2.78 &	90.19& 1.30& 0.34&2.65 &3.49 &45 &94  \\ \cline{2-11}	
				&0.80&500	    &2.72 &	80.25& 1.68& 0.33&2.40 &3.81 &48 &97 \\ \hline 
				\hline
				\multicolumn{11}{|c|}{Deviation Information-based Criterion by G. Zhang}\\ \hline
				Data &$1-\alpha$    &$N_{Trial}$    &$\mu_{PINAW}$    &	$\mu_{PICP}$ 	& $\sigma_{PICP}$   & $\mu_{PINAFD}$ & $\mu_{CWC}$ & $\mu_{CWFDC}$ & $\tilde{N}_{iter}^{(PICP(1\%))}$ & $\tilde{N}_{iter}^{(PINAW(1.5))}$\\ \hline
				\multirow{ 3}{*}{\rotatebox[origin=c]{90}{\parbox[c]{1cm}{\centering Wind Power}}}
				&0.95&500	    &5.23		&	94.15   & 3.42    & 0.89 &5.25 &8.62 &110 & 178   \\ \cline{2-11}
				&0.90&500	    &4.82		&	90.11   & 4.61    & 1.02 &5.83 &13.94 &102 & 159   \\ \cline{2-11}	
				&0.80&500	    &4.14		&	79.53   & 6.89    & 1.50 &8.12 &22.5 &106 & 162 	 \\ \hline 
				\multirow{ 3}{*}{\rotatebox[origin=c]{90}{\parbox[c]{1cm}{\centering E. De-mand}}}
				&0.95&500	    &2.10 &	95.06 &2.09& 0.28 &2.20 &2.71 &78 &152   \\ \cline{2-11}
				&0.90&500	    &2.05 &	90.09 &2.67& 0.22 &2.09 &2.99 &83 &155  \\ \cline{2-11}	
				&0.80&500	    &1.93 &	80.01 &3.49& 0.20 &2.01 &4.43 &89 &174	 \\ \hline 
				\hline
				\multicolumn{11}{|c|}{Proposed Method ($\rho$ =1, $\beta$ =1000, $\delta$ = $\alpha$/50)}\\ \hline
				Data &$1-\alpha$    &$N_{Trial}$    &$\mu_{PINAW}$    &	$\mu_{PICP}$ 	& $\sigma_{PICP}$   & $\mu_{PINAFD}$ & $\mu_{CWC}$ & $\mu_{CWFDC}$ & $\tilde{N}_{iter}^{(PICP(1\%))}$ & $\tilde{N}_{iter}^{(PINAW(1.5))}$\\ \hline
				\multirow{ 3}{*}{\rotatebox[origin=c]{90}{\parbox[c]{1cm}{\centering Wind Power}}}
				&0.95&1000	    &7.40		&	95.09   & 0.11    & 0.53 &7.40 &8.03 &45 & 92   \\ \cline{2-11}
				&0.90&1000	    &6.01		&	90.18   & 0.32    & 0.58 &6.01 &9.99 &41 & 97   \\ \cline{2-11}	
				&0.80&1000	    &4.31		&	80.41   & 0.38    & 0.67 &4.31 &10.08 &42 & 88 	 \\ \hline
				\multirow{ 3}{*}{\rotatebox[origin=c]{90}{\parbox[c]{1cm}{\centering E. De-mand}}}
				&0.95&1000	    &2.34 &95.10 & 0.14 & 0.31 &2.36 &2.67 &43 &93   \\ \cline{2-11}
				&0.90&1000	    &2.29 &90.21 & 0.41 & 0.23 &2.35 &3.02 &40 &90  \\ \cline{2-11}	
				&0.80&1000	    &2.27 &80.40 & 0.59 & 0.20 &2.33 &3.48 &28 &84	 \\ \hline 
			
			\end{tabular} 
			\\
			$\mu_{CWC} = \mu_{PINAW} + \gamma^{(\alpha, PICP)} (\mu_{PICP})$; \ 
			$\mu_{CWFDC} = \mu_{PINAW} + \mu_{PINAFD} + 1000* (1-\alpha+\delta -\mu_{PICP})^2$\\
			$\tilde{N}_{iter}^{(PICP(1\%))}$ is median of the number of iterations to reach $|1-\alpha + \delta -PICP| < 1\%$\\
			$\tilde{N}_{iter}^{(PINAW(1.5))}$ is median of the number of iterations to reach $PINAW < 1.5 \times PINAW_{Opt}$\\
\end{table*}

\subsection{C. Wan's method}
Four neurons with different structures and random initialization are trained and the one corresponding to the lowest $\gamma |ACE| + \lambda |S_{AV}|$ value is selected. Fig. \ref{Optimal_c}(b) presents those lowest cost function values. The number of neurons is varied from 5 to 15. The optimal NN size is found to be 10, 9, and 9 for PINC = 95\%, 90\%, and 80\% respectively for the wind power data. Similarly, NN size is found to be 11, 8, and 7 for PINC = 95\%, 90\%, and 80\% respectively for the electricity demand data.

Five hundred NNs of optimal size are trained for both of the wind power and the electricity demand data. The reported result considers NNs which provide logical PIs. With the C. Wan's cost function, roughly 99\% NN training converges and all converged NNs provide a logical PI.  The performance of that method is presented as the second segment of Table \ref{TABComp1}.  C. Wan's method provides PIs of slightly higher width but reduces the average failure distance.

\subsection{L. G. Marn's method}
L. G. Marn's method considers the deviation from the mid-interval.
This time, NNs with the lowest $PINAW +  ||e||^2 /n$ values are considered. Fig. \ref{Optimal_c}(c) presents the lowest cost function values. Following the same process, optimal NN size is found to be 11, 10, and 10 for PINC = 95\%, 90\%, and 80\% respectively for the wind power data. NN size is found to be 11, 9, and 8 for PINC = 95\%, 90\%, and 80\% respectively for the electricity demand data.

As this cost function and its derivatives are continuous, the convergence is higher (99.2\%).  The performance of the method is presented as the third segment in Table \ref{TABComp1}.  As a slightly higher PICP compared to PINC is also penalized in this cost function, the average PICP becomes slightly greater compared to other methods. As Marn's method does not optimize the failure distance, the failure distance is greater compared to C. Wan's method. In the contrary, consideration of the deviation from the mid-interval results in a lower failure distance compared to the LUBE method. Also, optimization of the deviation from the mid-interval brings the most probable regions of targets close to the middle of PIs. That results in slightly wider PIs.


\begin{figure}
\begin{center}
\includegraphics[width=8.7cm]{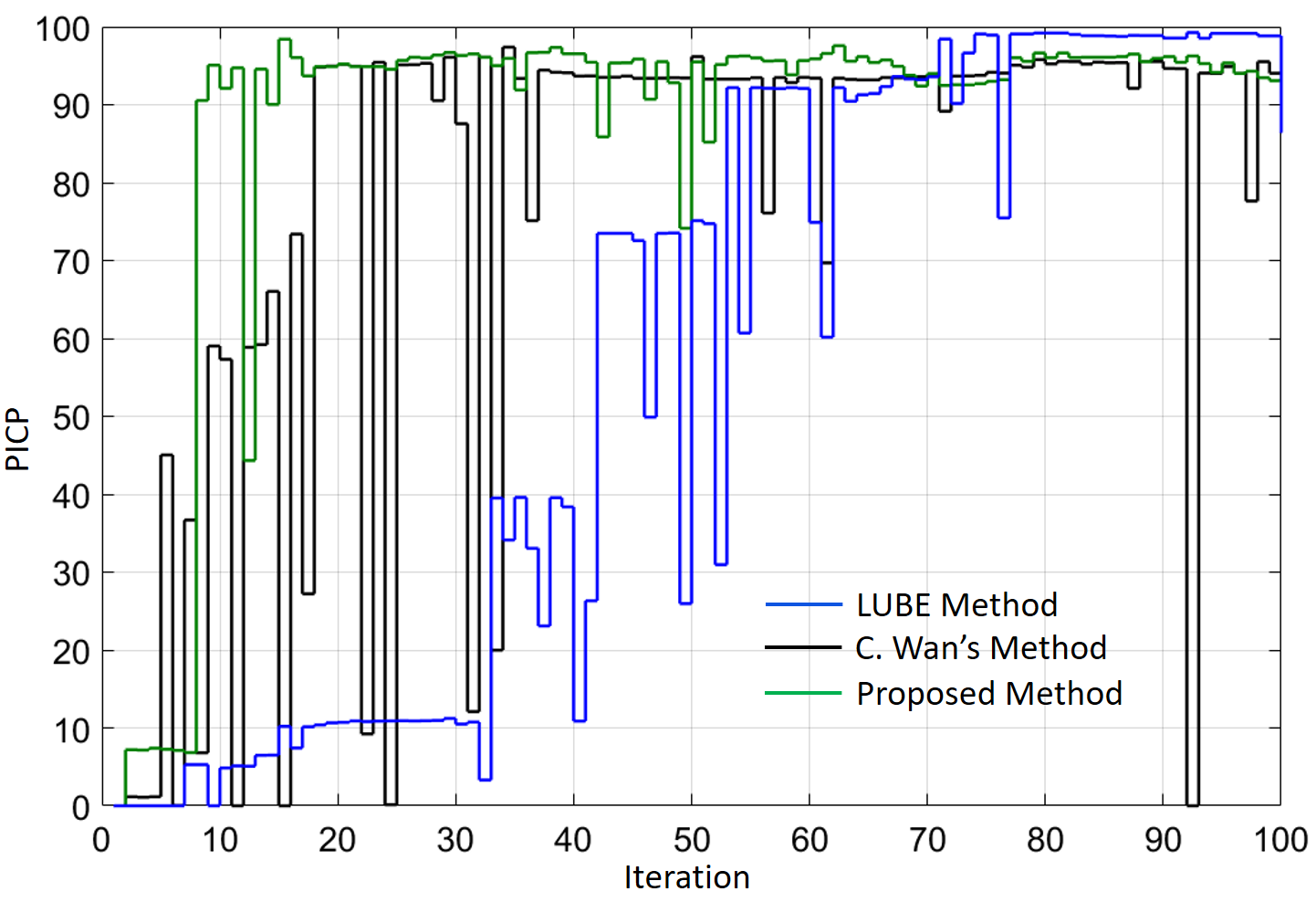}
\caption{\label{PICP_ITER} Typical representative PICP values over iterations for the proposed method and two existing methods. Only the first 100 iterations are presented. These values vary from simulation to simulation.}
\end{center}
\end{figure}

\begin{figure}
\begin{center}
\includegraphics[width=8.75cm]{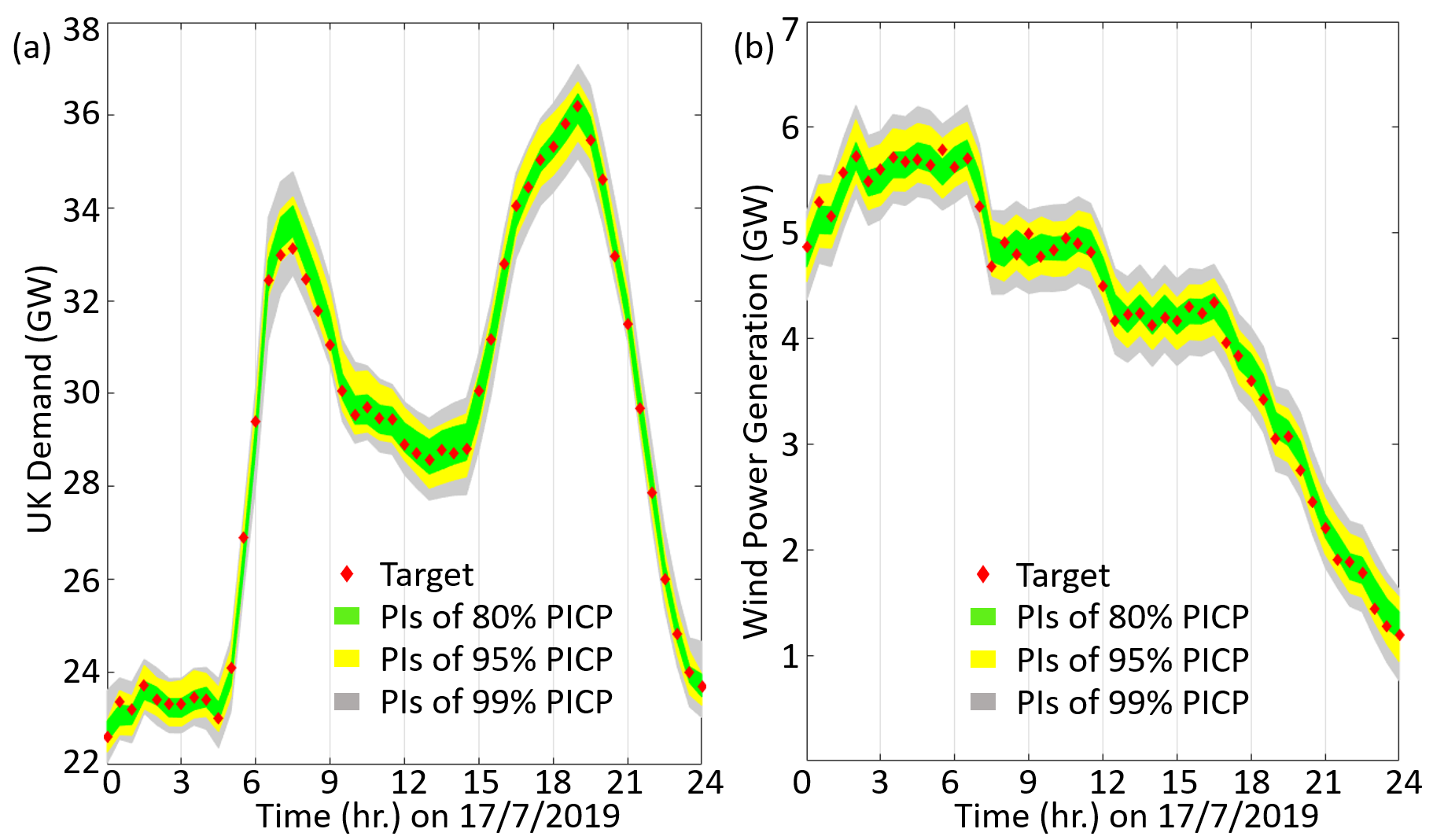}
\caption{\label{Optimal_PI_Wind} PIs of 80\%, 95\%, and 99\% PICP with targets for (a) the electricity demand and (b) wind power generation of UK grid on 17/7/2019. A 5-minutes ahead UQ is performed. Two samples are drawn in each hour for better visualization. Intervals are computed through an optimally trained NN. The structure of the NN is presented in Fig. \ref{Optimal_PI_NN}. Four previous samples and the time is used to compute the interval.}
\end{center}
\end{figure}

\begin{figure}
\begin{center}
\includegraphics[width=8.75cm]{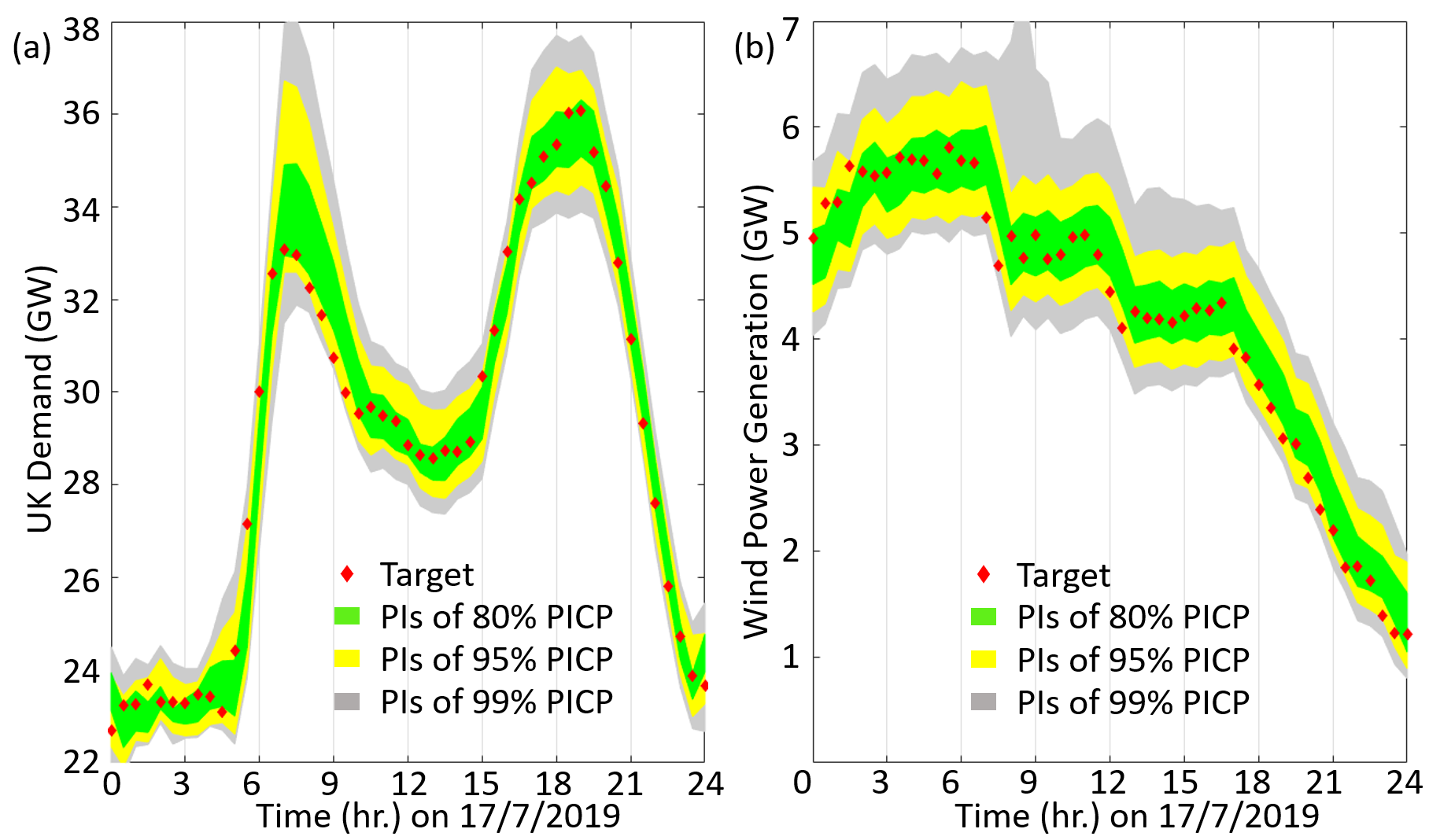}
\caption{\label{Optimal_PI_Wind30} PIs of 80\%, 95\%, and 99\% PICP with targets for (a) the electricity demand and (b) wind power generation of UK grid on 17/7/2019. A 30-minutes ahead UQ is performed. Intervals are wider compared to the 5-minutes ahead UQ.}
\end{center}
\end{figure}

\subsection{G. Zhang's method}
The same simulations are performed for NNs trained using G. Zhang's cost function. The lowest $DIC$ values are considered for drawing Fig. \ref{Optimal_c}(d). Optimal NN size is found to be 10, 9, and 9 for PINC = 95\%, 90\%, and 80\% respectively for the wind power data. Optimal NN size is 11, 9, and 8 for PINC = 95\%, 90\%, and 80\% respectively and for the electricity demand data.

Due to the discontinuity of the cost function at PICP=PINC, the convergence of that simulation is low (96.2\%).  The performance of that method is presented as the fourth segment in Table \ref{TABComp1}.  This function also provides a good PICP on average and a high failure distance. As the low PICP is poorly penalized and the sum of failure distance and the average width may potentially result in a wrong gradient direction, the variation in PICP is higher compared to other methods.



\subsection{The proposed method}
NNs are trained using the proposed cost function as a part of the performance evaluation. The lowest $PINAW + PINAFD$ values among four NNs are considered for drawing Fig. \ref{Optimal_c}(e). Optimal NN size is found to be 10 for all PINC values (95\%, 90\%, and 80\%) for the wind power generation. Optimal NN size is found to be 12, 10, and 9 respectively for 95\%, 90\%, and 80\% PINC for the electricity demand data. 

The failure distance resistance parameter ($\rho$) is set to one to provide an equal concentration towards width and the failure distance. The PI coverage penalty factor ($\beta$) is set to one thousand to reduce the PICP variation among different NNs. According to observations, the PICP variation is higher for a higher $\alpha$. Therefore, the coverage margin ($\delta$) is set to ($\alpha/50$). One thousand NNs of optimal size are trained and NNs providing logical PIs are considered. 99.8\% of NN training is converged and provides a logical PI. The probable reason for 0.02\% convergence failure can be PICP = 0 for the first thousand iterations. Although the NN training process changes weights in each iteration, the PICP remains zero and the change in PINAW + PINAFD is negligible compared to the large failure penalty at PICP = 0. Thus, the gradient becomes small and causing slow convergence. 

Although the simulation does not converge in 0.2\% situations, the convergence of the proposed system is much better compared to any existing method. The simulation does not converge in 0.8\% to 9\% situations in currently available methods. Fig. \ref{PICP_ITER} presents PICP values over iterations. The value of PICP reaches a feasible range (PINC - $\alpha$ $<$ PICP $<$ 1) within the first thirty iterations most of the time in the proposed method. Usually, more than fifty iterations are required to get such a feasible range in the initial LUBE method. Other continuous cost functions also require more than thirty iterations to get such a feasible range. Moreover, the PICP iterates near 100\% when a greater PICP (PICP $>$ PINC + $\delta$) is not penalized. Variation in the finalized PICP is higher in LUBE and other continuous cost functions due to discontinuity at the gradient near PICP = PINC. As the proposed cost function is continuous along with its high order derivatives, the convergence is faster and possesses a low variance.

The performance of this method is presented as the last segment of Table \ref{TABComp1}.  PICP variation becomes much lower with the proposed method when $\beta$ is set to 1000. The mean value of PICP remains very close to PINC + $\delta$ and the PICP variation is close to $\delta$. As a result, about 85\% of NNs maintains PICP $>$ PINC. On other systems, less than 70\% of NNs maintains PICP $>$ PINC. The proposed system has much lower PINAW + PINAFD as it is the performance criterion during PICP = PINC + $\delta$. However, the user may choose any desired failure distance resistance parameter ($\rho$) according to their preference. Table \ref{TABComp1} also present the convergence of the neural network training. Both PICP and PINAW reaches near to their optimum value with fewer iterations with the proposed cost function. A few high values can change the average greatly, therefore we consider median values to compare convergences.

Fig. \ref{Optimal_PI_Wind} presents PIs with targets for the 5-minutes-ahead prediction with the proposed NN training. Fig. \ref{Optimal_PI_Wind}(a) presents PIs of three different PICPs with targets for the electricity demand data.  Fig. \ref{Optimal_PI_Wind}(b) presents PIs of three different PICPs with targets for the wind power generation data. PIs and corresponding targets are drawn on the same plot to visualize the PI of the heteroscedastic system.

Fig. \ref{Optimal_PI_Wind30} presents PIs with targets for the 30-minutes-ahead prediction with the proposed NN training. Fig. \ref{Optimal_PI_Wind30}(a) presents PIs of three different PICPs with targets for the electricity demand data.  Fig. \ref{Optimal_PI_Wind30}(b) presents PIs of three different PICPs with targets for the wind power generation data. 









\section{Conclusion}
Uncertainty is inescapable but uncertainty aware decisions bring higher sustainability and profitability. The NN based LUBE PI construction method achieved state of the art performance in quantifying asymmetrically heteroscedastic uncertainty in terms of narrow width and required PICP. However, NNs need to be re-trained for new types of signals and the non-smooth LUBE cost function often fails to achieve an efficient uncertainty guided NN. Several improvements to the LUBE method have proposed by different researchers for optimal NN training. Each one of them has some limitations in terms of convergence, understandability, smoothness, parameter insufficiency, or customizability. Therefore, a smooth optimization function is proposed. A low failure distance results in a low non-coverage penalty. Moreover, the user may prefer to minimize the failure distance. Therefore, the cost function considers coverage, width, and failure distance criteria to train NNs for the construction of PIs with higher consistency. Researchers may bring 100\% convergence of the training in the future.







\bibliographystyle{elsarticle-num}
\bibliography{References_PI.bib}

\end{document}